\documentclass[journal,transmag]{IEEEtran}
\usepackage{graphicx}
\usepackage{graphics}
\usepackage{subfigure}
\usepackage{calligra}

\usepackage{mathalfa}
\usepackage{amsmath}
\usepackage{amsfonts}
\usepackage{amssymb}
\usepackage{amsmath}
\usepackage{fancyhdr}

\ifCLASSINFOpdf
\else
\fi
\begin{document}
\title{Gauge theory and twins paradox of disentangled representations}


%Montgomery Scott\IEEEauthorrefmark{3}, and
%Eldon Tyrell\IEEEauthorrefmark{4},~\IEEEmembership{Fellow,~IEEE}}
%\IEEEauthorblockA{\IEEEauthorrefmark{1}School of Electrical and Computer Engineering,
%Georgia Institute of Technology, Atlanta, GA 30332 USA}
%\IEEEauthorblockA{\IEEEauthorrefmark{2}Twentieth Century Fox, Springfield, USA}
%\IEEEauthorblockA{\IEEEauthorrefmark{3}Starfleet Academy, San Francisco, CA 96678 USA}
%\IEEEauthorblockA{\IEEEauthorrefmark{4}Tyrell Inc., 123 Replicant Street, Los Angeles, CA 90210 USA}% <-this % stops an unwanted space
%\thanks{Manuscript received December 1, 2012; revised August 26, 2015.
%Corresponding author: M. Shell (email: http://www.michaelshell.org/contact.html).}}

\author{\IEEEauthorblockN{Xiao Dong, Ling Zhou}
\IEEEauthorblockA{Faculty of Computer Science and Engineering, Southeast University, Nanjing, China}}

%{Shell \MakeLowercase{\textit{et al.}}: Bare Demo of IEEEtran.cls for IEEE Transactions on Magnetics Journals}



% for Transactions on Magnetics papers, we must declare the abstract and
% index terms PRIOR to the title within the \IEEEtitleabstractindextext
% IEEEtran command as these need to go into the title area created by
\IEEEtitleabstractindextext{%
\begin{abstract}
Achieving disentangled representations of information is one of the key goals of deep network based machine learning system. Recently there are more discussions on this issue. In this paper, by comparing the geometric structure of disentangled representation and the geometry of the evolution of mixed states in quantum mechanics, we give a fibre bundle based geometric picture of disentangled representation which can be regarded as a kind of gauge theory. From this perspective we can build a connection between the disentangled representations and the twins paradox in relativity. This can help to clarify some problems about disentangled representation.
\end{abstract}

}

\maketitle

%\affiliation{}

%\tableofcontents

\IEEEdisplaynontitleabstractindextext

% \fi
\IEEEpeerreviewmaketitle

\section{Introduction}

Achieving a disentangled representation of data is a key issue in machine learning systems. But what's the essence of disentangled representations? How can we define it? Can we guarantee to reach disentangled representations in machine learning systems? In a recent paper \cite{Locatello2018Challenging} it was claimed that unsupervised learning of disentangled representations is fundamentally impossible without inductive biases on both the model and the data. This work advocated lots of discussions on disentangled representations.

But till today even a commonly accepted formal definition of disentangled representation is not available yet. We attribute this situation to that we are still lacking of an understanding of the interpretability of deep learning systems. A fully understanding of disentangled representations is only available with a clear interpretability of deep networks. This is to say, we need to examine the disentangled representation problem in a mathematical framework of deep learning systems so that we can define and analysis the properties of disentangled representations.

In this paper we will try to formulate disentangled representations from a geometric point of view following the programme of geometrization of deep networks\cite{Dong2019geo}\cite{Dong_deep}\cite{Dong2019over}, which was proposed as a framework for the interpretability of deep learning systems. The key idea of this geometrization programme is that, deep networks are  physical so that deep networks and physics should have the same mathematical picture. Therefore the well known geometrization of physics can be adapted in the field deep learning as a framework to understand deep learning systems. This idea has shown some promising results by finding the correspondence between deep learning system and physics, for example we find different geometric structures in deep networks which correspond to the geometric structures in general relativity, quantum measurement, quantum computation and quantum gravity.

To formulate disentangled representations, we will explore a new geometric structure, a fibre bundle structure. We will see disentangled representations can be understood as a gauge theory which has a counterpart in the evolution of mixed quantum states. By comparing the geometric structures of disentangled representations and the mixed quantum states, we can give a geometric picture for disentangled representations. We will also use this perspective to give a negative argument to the conclusion of \cite{Locatello2018Challenging}.

\section{Geometrization of deep networks}
Geometrization of deep networks aims to formulate deep networks and deep learning systems by geometrizaiton. That's to say deep networks can be described by geometric structures and the properties of deep learning systems are determined by their correspondent geometric pictures. The underlining reason for us to propose this idea is that we believe deep networks are physical in the meaning that deep networks are efficient descriptors of our physical world and our physical world is generated by deep networks, to be more specific, by deep networks of quantum computations according to computational university. So the geometrization of physics, which is one of the most fundamental thought in physics, stems from the geometrization of deep networks. So the global picture of geometrization of deep networks is that, we can derive our physical world with all its rules from deep networks. On the other hand, ideas from physics research can be borrowed for the interpretability of deep networks.

Currently the main components of geometrization of deep networks include:\\
(1) The rule of deep networks can be formulated as a least action principle.\\
(2) The action here is defined by an information metric based complexity of deep networks.\\
(3) The basic geometric framework of deep systems is the Riemannian geometry. We point out here this is compatible with the idea that deep networks can be understood as differential equations or optimal control problems since they can also be formulated on Riemannian geometry.\\
(4) There exist a dictionary between the concepts of deep networks and physics, just as the dictionary between physics and geometry.\\
(5) Important properties of deep learning systems such as convergence, generalization, disentanged representation, network architecture search, knowledge distillation and network pruning, can all be understood geometrically.\\
(6) The structure and laws of our physical world can be derived from deep networks so that we also have a better interpretability of physics. For example now we already have a rough idea how spacetime, material and their interactions can be understood as emergent phenomena from the deep network of quantum computation.\\

In the remaining part of this paper, we will follow the aforementioned road map to formulate a geometric picture of disentangled representations, which turns out to be a fibre bundle structure and it can be regarded as a gauge theory of disentangled representations.

\section{Geometry of disentangled representations}

\subsection{Geometry of the evolution of mixed quantum states}
To give the geometry of disentangled representation, we need first understand the geometry of the evolution of mixed quantum, which is an analogue of the geometry of disentangled representations.

We have two types geometric structures for the evolution of mixed quantum states as in \cite{Heydari_dynamicdiatance}\cite{Heydari_geometryquantum}\cite{Montgomery1991Heisenberg}. It can be proved that they are essentially the same. So our formulation to this geometric structure will be an integrated version of these two structures.

Mixed quantum states can be represented by density operators denoted by $\rho$, which are self-adjoint nonnegative unit trace operators. The space of density operators of a quantum system with a Hilbert space $\mathcal{H}$ is denoted by $D(\mathcal{H})$. A density operator evolves according to a von Neumann equation and remains in an orbit of the left conjugation action of the unitary group of $\mathcal{H}$. There exists a one-to-one correspondence between the orbits and the spectra of $\rho$ with $k$ non-increasing nonnegative eigen values $\sigma={p_1,p_2,...,p_k}$. We denote the orbit corresponding to the spectrum $\sigma$ as $D(\sigma)$.

Assuming $\rho_0,\rho_1$ are two isospectral density operators on $D(\sigma)$ and $H$ is a Hamiltonian operator on $\mathcal{H}$ that evolves $\rho_0$ to $\rho_1$ with an evolutional trajectory curve $\rho$, which satisfies:
\begin{equation}\label{eq-1}
\begin{split}
i\dot{\rho}&=[H,\rho]\\
\rho(t_0)&=\rho_0\\
\rho(t_1)&=\rho_1\\
\end{split}
\end{equation}

We can define the H-distance between $\rho_0,\rho_1$ to be the path integral of the uncertainty of $H$ along $\rho$ given by
\begin{equation}\label{eq-2}
  D_H(\rho_0,\rho_1)=\int_{t_0}^{t_1}\sqrt{Tr(H^2\rho)-Tr(H\rho)^2}dt
\end{equation}
Physically the distance is just the energy needed to evolve $\rho_0$ to $\rho_1$ using $H$. We can then define the dynamic distance between $\rho_0$ and $\rho_1$ as
\begin{equation}\label{eq-3}
  D(\rho_0,\rho_1)=\min_{H}D_H(\rho_0,\rho_1)
\end{equation}
The dynamic distance is a proper distance among isospectral states, which is also unitary invariant so that
\begin{equation}\label{eq-4}
  D(U\rho_0 U^\dagger,U\rho_1 U^\dagger)=D(\rho_0,\rho_1)
\end{equation}

Standard purification of $D(\mathcal{H})$ is denoted by $\Psi \in \mathcal{S}(\mathcal{H}\otimes C^{k*})$ and the surjective map $\pi: \mathcal{S}(\mathcal{H}\otimes C^{k*})\rightarrow D, \pi(\Psi)=\Psi\Psi^\dagger$. Let $P(\sigma)=Diag(\sigma)$ taking $\sigma$ as its diagonal and $\mathcal{S}(\sigma)$ be the subspace of those $\Psi$ that satisfies $P(\sigma)=\Psi^\dagger\Psi$, then $\mathcal{S}(\sigma)$ is a principle fibre bundle over $\mathcal{D}(\sigma)$ with a right acting gauge group $\mathcal{U}(\sigma)$ so that for $u \in \mathcal{U}(\sigma)$, we have

\begin{equation}\label{eq-5}
\begin{split}
\Psi\Psi^\dagger &= \rho \\
\Psi^\dagger\Psi &= P(\sigma) \\
u\cdot \Psi &= \Psi u \\
uP(\sigma)&=P(\sigma)u
\end{split}
\end{equation}

We can then define the vertical and horizontal bundles as $V\mathcal{S}(\sigma)=Ker d\pi$ and $H\mathcal{S}(\sigma)=V\mathcal{S}(\sigma)^\perp$. Vectors in $H\mathcal{S}(\sigma)$ and $V\mathcal{S}(\sigma)$ are called
horizontal and vertical vectors respectively. A curve in $\mathcal{S}(\sigma)$ is called horizontal if is velocity vectors are all horizontal. For every curve $\rho$ in $\mathcal{D}(\sigma)$ and $\Psi_0$ in the fibre over the initial state $\rho_0$, there is a unique horizontal lift of $\rho$ in $\mathcal{S}(\sigma)$ starting from $\Psi_0$. The horizontal and vertical bundles define a connection on $\mathcal{S}(\sigma)$.

Now we have two Riemannian structures. One the Riemannian structure on $\mathcal{D}(\sigma)$ which defines the dynamical distance between isospectral density operators as the minimal energy needed to evolve $\rho_0$ to $\rho_1$. If the evolution process is regarded as a quantum computation process, the dynamical distance is just the minimal computational complexity to achieve the quantum algorithm $U$ that satisfies $\rho_1=U\rho_0U^\dagger$.

\begin{figure}
  \centering
  \includegraphics[width=8.5cm]{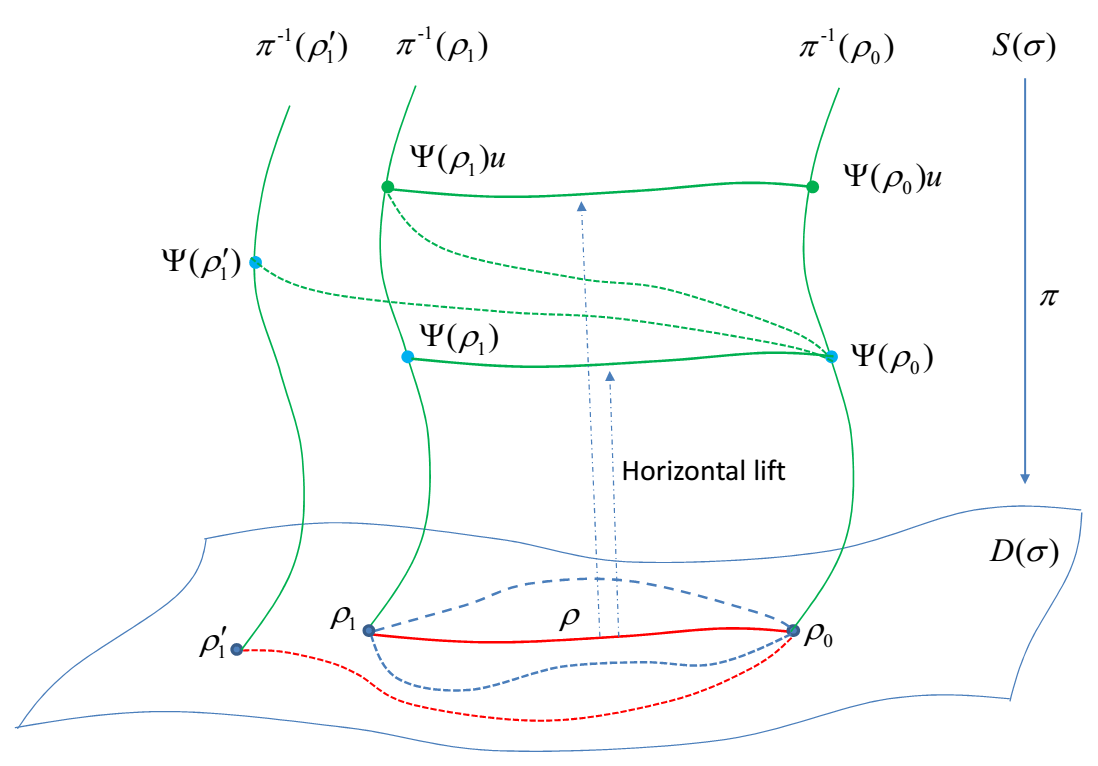}
  \caption{Geometric structure of the evolution of mixed quantum states. $\rho_0$,$\rho_1$ and $\rho_1^{'}$ are all isospectral density operators on $\mathcal{D}(\sigma)$. The different curves on $\mathcal{D}(\sigma)$ are different evolution trajectories between $\rho_0$ and $\rho_1$, or different quantum algorithms to transform $\rho_0$ to $\rho_1$. The geodesic $\rho$ is the optimal algorithm with a computational complexity that equals to the dynamical distance between $\rho_0$ and $\rho_1$. Points on the fibre $\pi^{-1}(\rho_0)$ and $\pi^{-1}(\rho_1)$ are different purification of $\rho_0$ and $\rho_1$ respectively. The two horizontally lifted geodesics in $\mathcal{S}(\sigma)$ of the geodesic $\rho$ are physically equivalent.} \label{Fig-1}
\end{figure}

The other geometric structure is the Riemannin structure on $\mathcal{S}(\sigma)$ defined by the connection on it. Accordingly for two isospectral density operators $\rho_0,\rho_1$ we have geodesics between them on $\mathcal{D}(\sigma)$ and geodesics between points on the fibres $\pi^{-1}(\rho_0)$ and $\pi^{-1}(\rho_1)$. About the relationship between these two Riemannian structures, we only point out that the horizontal lift of a geodesic in $\mathcal{D}(\sigma)$ to $\mathcal{S}(\sigma)$ is also a geodesic in $\mathcal{S}(\sigma)$ and they have the same length. The fibre bundle structure is illustrated in Fig. \ref{Fig-1}.

In order to find the relationship with disentangled representations in deep networks, here we also give an explicit construction of the gauge field $\mathcal{U}(\sigma)$. In \cite{Montgomery1991Heisenberg} another fibre bundle of mixed states, the generalized Hopf fibration, was introduced where the gauge group is a left acting group $\mathcal{V}(\sigma)$ with $v \in \mathcal{V}(\sigma)$ satisfies
\begin{equation}\label{eq-6}
\begin{split}
v\cdot \Psi &= v\Psi  \\
v\rho_0 &=\rho_0 v
\end{split}
\end{equation}

In \cite{Montgomery1991Heisenberg} it was shown that if $\rho_0=U_0 Diag(\sigma) U_0^\dagger$ with $m_i$ be the multiplicites of the eigenvalues of $\rho_0$, then the gauge group is given by
\begin{equation}\label{eq-7}
  V(\sigma)=Ad_{U_0}\otimes_i U(m_i)
\end{equation}

So the gauge group $U(\sigma)$ is in fact just $\otimes_i U(m_i)$. This means the function of the gauge group is just to \emph{mix} the eigenvectors of $\rho_0$ with the same eigen value by an unitary operator. This in fact can be understood to \emph{entangle} those eigenvectors with the same eigenvalue.

From a computation point of view, $\rho_0,\rho_1$ are the input and output data of a computation process and the computation process is given by $\rho$. The geodesic on $\mathcal{D}(\sigma)$, which defines the dynamical distance between $\rho_0$ and $\rho_1$, is the optimal algorithm to accomplish the computation. The fibres $\pi^{-1}(\rho_0)$, $\pi^{-1}(\rho_0)$ are different ways to represent the input and output data. Curves on the fibre bundle to connect points in $\pi^{-1}(\rho_0)$ and $\pi^{-1}(\rho_0)$ are different implementations of the algorithm $\rho$.

According to the basic idea of gauge theory, different gauges are just different equivalent descriptions of the same physics. All the horizontal geodesics, the two horizontally lifted geodesics as shown in Fig. \ref{Fig-1} , on the fibre bundle are physically equivalent realizations of the optimal algorithm and they have exactly the same physical meaning.

\subsection{Geometric picture of disentangled representations}
Given the geometric picture of the evolution of mixed quantum states, which can be regarded as a geometric description of the most fundamental computation system, now we can scratch the geometric picture of disentangled representations of deep network.

The first concern about disentangled representations is, how to define disentangled representations? What's the fundamental nature of disentangled representations? Usually we think a disentangled representation should have the property that each data generative factor should be represented by a single or a small set of dimensions. There is also effort to define disentangled representations from the transformation properties of our world\cite{Higgins2018Disentangled}. The transformations that change only a subset of properties of the underlying world state while keeping all other properties invariant are regarded as disentangled representations of data. We can see here the word \emph{transformation} is in fact just an information processing process or a computation algorithm. So is it possible to derive the concept of disentangled representation purely from a computation point of view?

Our suggestion is to put the disentangled representation and the computational complexity on the same foot. A disentangled representation of certain data is just the optimal algorithm (with the minimal computational complexity) to generate that data. Or in other words, a disentangled representation is just the natural by-product of an optimization on the computational complexity. For example, in the aforementioned computational complexity of the evolution of mixed quantum states, the optimal algorithm to generate $\rho_1$ from $\rho_0$ is a Hamiltonian evolution of the quantum state with the minimal energy expense. From the quantum computational complexity or the Hamiltonian complexity point of view, the quantum circuit to achieve the optimal quantum algorithm consists of a sequence of simple quantum gates so that at each stage of the computation procedure, we only \emph{entangle} information from a few quantum bits. In quantum computation, the complexity of a pure quantum state $|\psi\rangle $is the minimal complexity of all quantum algorithms that can generate $|\psi\rangle$ from a product state $|00...0\rangle$. The optimal algorithm is in fact the optimal disentangled representation of the state $|\psi\rangle$.

From these perspective, in deep network based information processing systems, we can define the optimal deep network as the deep network with the minimal network complexity, where the network complexity is defined by the Fisher information metric. If the deep network is a generative model of data, then naturally it's also the optimal disentangled representation of data. Now we have an universal definition of disentangled representations and the essence of disentangled representations is just low computational complexity.

In our former work, we have shown that the complexity of deep networks has the same Riemannian structure as the complexity of quantum computation, which is just the Riemannian structure on $\mathcal{D}(\sigma)$ as given above. Now we would like to construct the fibre bundle structure of disentangled representations of deep networks corresponding to the fibre bundle structure of the evolution of mixed quantum state shown in Fig. \ref{Fig-1}.
    
Since this work aims to challenge the argument proposed in \cite{Locatello2018Challenging} that disentangled representations can not be obtained in unsupervised learning systems without inductive bias, we here construct the fibre bundle structure of disentangled representations directly based on the formulation of \cite{Locatello2018Challenging}.

In \cite{Locatello2018Challenging} their key argument is the Theorem 1, where the formulation of disentangled representations is as follows.
(1) The real world data $\mathbf{x}$ is generated by a two-step generative process. The first step is to generate a multivariate latent random variable $\mathbf{z}$ from a distribution $P(\mathbf{z})$. In the second step the observed data $\mathbf{x}$ is sampled from a conditional distribution $P(\mathbf{x}|\mathbf{z})$, where $P(\mathbf{x}|\mathbf{z})$ is achieved by a deep network. The goal of representations learning is to find a representation $r(\mathbf{x})$ of $\mathbf{x}$ that help us to better understand the structure of $\mathbf{x}$.\\

(2) Theorem 1 of \cite{Locatello2018Challenging}. For $d>1$ and $p(\mathbf{z})=\prod_{i=1}^d p(\mathbf{z}_i)$, there exists an infinite family of bijective functions $f$ so that $\mathbf{z}'=f(\mathbf{z})$ and $p(\mathbf{z})=p(\mathbf{z}')$, where $\mathbf{z}'$ is obtained by a complete entanglement of $\mathbf{z}$.

(3) \cite{Locatello2018Challenging} then argued that since the generative model $P(\mathbf{x}|\mathbf{z})$ can generate data $\mathbf{x}$ from both $\mathbf{z}$ and $\mathbf{z}'$ but $\mathbf{z}'$ is a fully entangled version of $\mathbf{z}$, so there is no way to distinguish the two generative models so that we have no way to define a disentangled representation of $\mathbf{x}$.
 
The fibre bundle structure of disentangled representations can be constructed by establishing a direct correspondence with the evolution of mixed states as follows:\\
(1) Since the generative model is only interested in distributions of $\mathbf{z},\mathbf{x}$, we can set $p(\mathbf{x})$ and $p(\mathbf{z})$ correspond to $\rho_0$ and $\rho_1$ respectively. We may have different distributions of $\mathbf{z}$ because we can have different representations of $\mathbf{x}$, but we have only a fixed $p(\mathbf{x})$.\\
(2) For a given $p(\mathbf{x})$ or $p(\mathbf{z})$, the correspondent fibre bundle of them are just the different data $\mathbf{x}$ and $\mathbf{z}$ that have the same distributions $p(\mathbf{x})$ or $p(\mathbf{z})$. So the base space of the fibre bundle is the distribution space of data and the total space is the data itself. For example, for a given distribution $p(\mathbf{z})$ in base space, its fibre bundle is exactly the data $\mathbf{z}'$ that is obtained by  $\mathbf{z}'=f(\mathbf{z})$, where $f$ is a function given in Theorem 1. For data $\mathbf{x}$, we have a similar structure where we denote the transformation function as $g$ instead of $f$.
(3) Following the idea of the computational complexity of quantum computation or the network complexity of deep networks, we have an optimal generative deep network $G$ that can generate the distribution $p(\mathbf{x})$ from $p(\mathbf{z})$. It should be point out that since the generative deep network to accomplish $P(\mathbf{x}|\mathbf{z})$ can only accomplish the task by transforming data samples of $p(z)$ to data samples of $p(x)$, the generative deep network is indeed working on the fibre bundle.

\begin{figure}
  \centering
  \includegraphics[width=8.5cm]{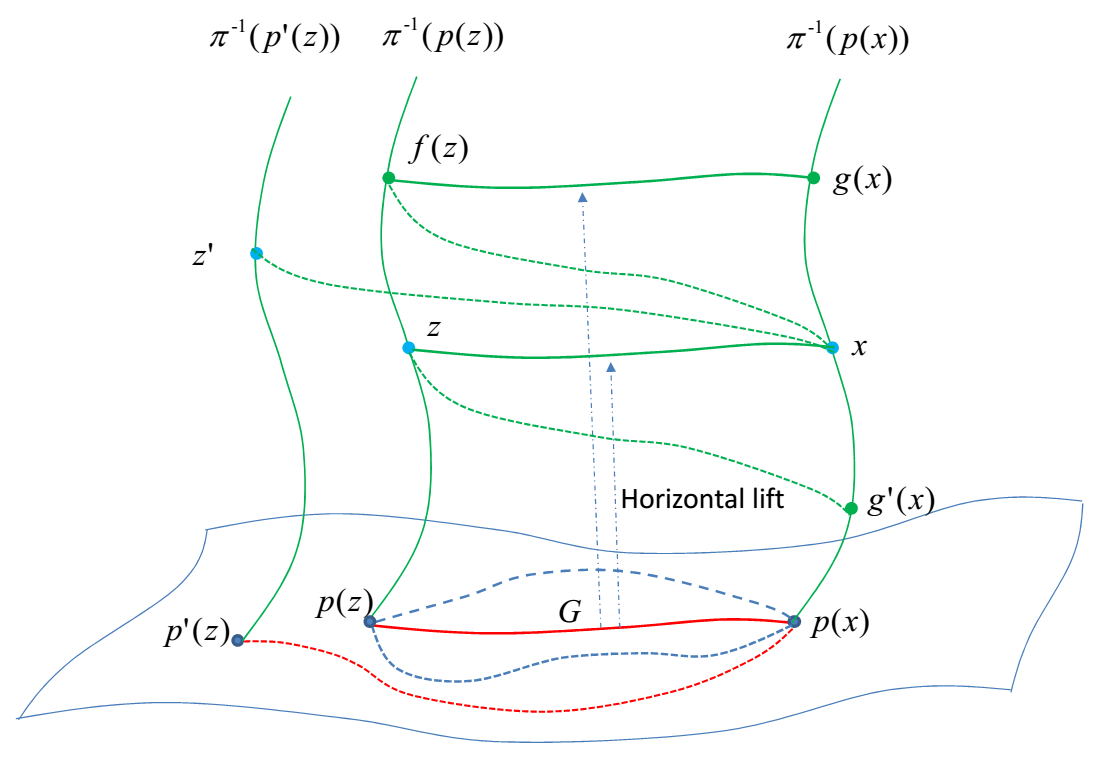}
  \caption{Geometric structure of disentangled representations. If $\mathbf{z}$ is the potentially optimal disentangled representation of data $\mathbf{x}$, the curve connection $\mathbf{x}$ and $\mathbf{z}$ represents the bidirectional optimal encoder/decoder. The two-step curve connecting $\mathbf{x}$,$f(z)$ and $g(x)$ is the encoder/decoder that encodes $\mathbf{x}$ as $\mathbf{z}'=f(\mathbf{z})$ and then decodes (generates) data $g(\mathbf{x})$ from $\mathbf{z}'$, where $g(\mathbf{x})$ has the same distribution of $\mathbf{x}$. Or equivalently the two-step curve connecting $\mathbf{g'(x)}$,$z$ and $x$ represents the same encoder/decoder procedure. These two pictures differ at if we choose the input of the encoder or the output of the decoder as our reference.} \label{Fig-2}
\end{figure}

Given this construction, the geometric structure of disentangled representations can be scratched as in Fig. \ref{Fig-2}. 
 
 The main argument of \cite{Locatello2018Challenging} is that since we can generate data $\mathbf{x}$, to be more exact the distribution of $\mathbf{x}$, from two entangled data $\mathbf{z}$ and $\mathbf{z}'$ with the same generative model, so we can not have an absolute criteria to judge which representation, $\mathbf{z}$ or $\mathbf{z}'$, is disentangled. This puzzling situation gives a contradictive picture about disentangled representations.

What kind of new understanding this geometric picture can bring us about disentangled representations? Are the two representations equivalent and indistinguishable? Is $\mathbf{x}$ entangled with $\mathbf{z}$ or $\mathbf{z}'$? We think we are facing a similar problem as the famous twins paradox in relativity. The paradox stems from the fact that when we talk about representation of data we only care about distributions of data, i.e. the base space of the fibre bundle, but the encoder/decoder works on data itself, i.e. the fibre bundle or the total space. 

(1)Are the two representations $\mathbf{z}$ and $\mathbf{z}'$ identical? The answer really depends on if we only consider the generator/decoder or the complete encoder/decoder process. If we only consider the generator or the decoder, both the representations $\mathbf{z}$ and $\mathbf{z}'$ give the same horizontally lifted curve as shown in Fig. \ref{Fig-2}. So physically they are identical according to the idea of gauge theory. But if we also consider the encoder, then obviously they are not equivalent since they correspond to different curves. So both  $\mathbf{z}$ and $\mathbf{z}'$ are disentangled representations for the output data of the generative model, but $\mathbf{z}$ is a disentangled representation of the input data of the encoder, while $\mathbf{z}'$ is an entangled description of the input data of the encoder. So they are not identical just as the twins are not in identical situation in the twin paradox. 

(2)Can we discriminate $\mathbf{z}$ or $\mathbf{z}'$? Yes. Intuitively from Fig. \cite{},  $\mathbf{z}$ and $\mathbf{z}'$ lead to different encoder/decoder curves so that by checking the complete circle of encoder/decoder, we can find the difference of $\mathbf{z}$ and $\mathbf{z}'$. Basically they have different computational complexities. 

(3)Which representation is the disentangled representation of $\mathbf{x}$, $\mathbf{z}$ or $\mathbf{z}'$? This is exactly the same question as in the twins paradox, which one of the twin brothers is younger? Of course we have an absolute answer, the one experienced acceleration is younger. In fact, the problem of time has a very close relationship with deep networks and the fundamental reason that the twin brothers experienced different time is that their trajectories have different computational complexities. Here we face the same situation, the encoder/decoder procedures of $\mathbf{z}$ and $\mathbf{z}'$ have different computational complexities. According to our definition of disentangled representations based on computational complexity, the one with the smaller complexity is a better disentangled representation. In this case, the disentangled representation is $\mathbf{z}$.

\section{Conclusions}
Geometrization of deep networks is a programme we proposed for the interpretability of deep networks and deep learning systems. By bridging physics and deep networks it aims to bring new interpretations for both fields. In this work, by comparing the fibre bundle structure of the evolution of mixed quantum states, we constructed the fibre bundle structure of disentangled representations, which can be regarded as a gauge theory of deep networks. This geometric formulation help us to propose a computational complexity based definition of disentangled representations and connect a recent puzzle about disentangled representations with the twins paradox in relativity. We hope this new geometric picture can be integrated in the programme of geometrization of deep networks and helps to build more connections between deep networks and physics. Another observation is that computational complexity plays a key role in both the fibre bundle structure and the concept of disentangled representations. In fact computational complexity of deep networks is also related with other key characteristics of deep learning systems such as the convergence, generalization etc. In physics, complexity is also becoming a central concept in quantum information processing, quantum gravity and quantum phase transition. We believe this is a strong sign that deep network and physics are essentially the same thing. We hope this idea can be accepted by more researchers in both fields.

\bibliographystyle{unsrt}

\bibliography{gauge}



%\vfill


\end{document}